\documentclass[letterpaper, 10 pt, conference]{ieeeconf}  

\usepackage{amsmath} 

\IEEEoverridecommandlockouts                              

\usepackage{amsmath}
\usepackage{amssymb}
\usepackage{booktabs}   
\usepackage{multirow}   
\usepackage{makecell}   
\usepackage{graphicx}   
\usepackage{xcolor}     
\usepackage{booktabs}
\usepackage{multirow}
\usepackage{graphicx}
\usepackage{comment}
\usepackage{algorithm}
\usepackage{algorithmic}
\usepackage{booktabs}
\usepackage{multirow}
\usepackage{graphicx}

\title{\LARGE \bf
KnowDiffuser: A Knowledge-Guided Diffusion Planner with LLM Reasoning
}

\author{
Fan Ding$^{1}$,
Xuewen Luo$^{2}$,
Fengze Yang$^{2}$,
Bo Yu$^{2}$,
HwaHui Tew$^{1}$,
Ganesh Krishnasamy$^{1}$,
Junn Yong Loo$^{1}$%
\thanks{*Corresponding author.}%
\thanks{$^{1}$Monash University Malaysia.}%
\thanks{$^{2}$University of Utah, USA.}
}

\begin{document}

\maketitle
\thispagestyle{empty}
\pagestyle{empty}

\begin{abstract}

Recent advancements in Language Models (LMs) have demonstrated strong semantic reasoning capabilities, enabling their application in high-level decision-making for autonomous driving (AD). However, LMs operate over discrete token spaces and lack the ability to generate continuous, physically feasible trajectories required for motion planning. Meanwhile, diffusion models have proven effective at generating reliable and dynamically consistent trajectories, but often lack semantic interpretability and alignment with scene-level understanding. To address these limitations, we propose \textbf{KnowDiffuser}, a knowledge-guided motion planning framework that tightly integrates the semantic understanding of language models with the generative power of diffusion models. The framework employs a language model to infer context-aware meta-actions from structured scene representations, which are then mapped to prior trajectories that anchor the subsequent denoising process. A two-stage truncated denoising mechanism refines these trajectories efficiently, preserving both semantic alignment and physical feasibility. Experiments on the nuPlan benchmark demonstrate that KnowDiffuser significantly outperforms existing planners in both open-loop and closed-loop evaluations, establishing a robust and interpretable framework that effectively bridges the semantic-to-physical gap in AD systems.

\end{abstract}

\section{Introduction}
The applications of Language Models (LMs) in the field of Autonomous Driving (AD) have attracted widespread attention due to their outstanding performance and emergent reasoning abilities \cite{cui2024survey}. Numerous recent studies have shown that LMs perform impressively across a variety of driving tasks \cite{luo2024pkrd}, such as DriveVLM \cite{tian2024drivevlm}, DriveGPT4 \cite{xu2024drivegpt4}, DiLu \cite{wen2023dilu}, and VLM-MPC \cite{long2024vlm}. By leveraging their powerful situational awareness, LMs enable more interpretable and flexible AD systems that can incorporate human-like reasoning, adapt to complex scenarios, and integrate commonsense knowledge into driving. Their emergence has opened up a promising new knowledge-driven paradigm for AD, pushing beyond the limitations of traditional rule-based and data-driven architectures \cite{lin2025causal}.

As a rising star in the AD domain, LMs have demonstrated remarkable capabilities. However, they struggle with directly generating precise low-level outputs such as trajectory points \cite{xing2025openemma}, which require fine-grained, physically feasible control. This limitation highlights the urgent need for a coordinated approach where knowledge-driven high-level decisions are effectively aligned with physically executable low-level trajectories, thereby enhancing the application of LMs in AD systems \cite{jiang2024senna, yang2023llm4drive}. To bridge this gap, it is essential to introduce models that are well-suited for generating continuous, mathematically accurate, and dynamically feasible trajectories. While traditional optimization-based or autoregressive approaches have been explored for trajectory generation, they often struggle to balance diversity, realism, and controllability in complex driving scenarios.

In contrast, diffusion models generate trajectories through a probabilistic denoising process that enables direct sampling of multi-modal and physically plausible outcomes from a Gaussian distribution \cite{sharan2024plan, liao2024diffusiondrive, zheng2025diffusion}. This formulation naturally captures environmental uncertainty and the diversity of driving behaviors \cite{chen2024overview}. Their iterative refinement ensures physical continuity and feasibility. Moreover, since the generative process is inherently guidable, external constraints such as LM commands can be seamlessly incorporated to steer generation toward task-aligned solutions \cite{lian2023llm, yang2024diffusion}. These properties make diffusion models particularly well-suited as controllable low-level generators in AD systems, complementing LM-based decision-making by bridging high-level semantic intent with executable motion plans. However, the iterative nature of the denoising process introduces relatively high inference latency, posing challenges for real-time deployment in time-critical AD settings \cite{mao2023leapfrog, li2025predictive}.

Therefore, we propose \textbf{KnowDiffuser}, a knowledge-guided diffusion planning framework that systematically bridges high-level semantic reasoning with low-level trajectory generation. The high-level module leverages an LM to interpret traffic contexts and produce knowledge-informed driving meta-actions that incorporate traffic rules, social norms, and human intent. To align these semantic decisions with physically grounded trajectories, we introduce a bridge mechanism that maps meta-actions to prior trajectories derived from historical driving data. Building on this, we design an efficient diffusion trajectory generator that integrates a truncated inference process with a two-step denoising strategy: instead of sampling from pure noise, the model initializes from semantically aligned priors with minimal noise injection, enabling fast and physically plausible trajectory generation that meets real-time planning demands.

Our contributions are summarized as follows:

\begin{itemize}
\item We propose KnowDiffuser, a novel motion planning framework that seamlessly integrates an LM with a diffusion-based trajectory generator. This hybrid design introduces a bridge mechanism that aligns high-level semantic decisions with low-level control trajectories, enabling interpretable and context-aware planning in complex driving scenarios.

\item We develop an efficient training and inference strategy for real-time diffusion-based trajectory generation. Specifically, we combine a truncated denoising process with a two-step refinement scheme, significantly reducing inference latency while preserving trajectory quality and feasibility.

\item We conduct extensive experiments on the nuPlan benchmark under open-loop and closed-loop settings. Results show that KnowDiffuser consistently outperforms state-of-the-art baselines in motion planning, demonstrating its effectiveness, efficiency, and potential for real-world deployment.
\end{itemize}

\section{Related Work}

\subsection{Diffusion Models for Trajectory Generation}

Diffusion models have emerged as a leading paradigm for trajectory generation in autonomous driving due to their ability to model complex, multimodal, and stochastic behaviors. By iteratively denoising Gaussian noise, these models produce physically feasible and diverse trajectories that capture the uncertainty inherent in dynamic traffic environments~\cite{ho2020denoising, sohl2015deep, song2021scorebased}. Early efforts in motion prediction, such as MID~\cite{gu2022stochastic}, Leapfrog~\cite{mao2023leapfrog}, and MotionDiffuser~\cite{jiang2023motiondiffuser}, demonstrate their strength in capturing multi-agent interactions and temporal variability. More recently, diffusion models have been extended to planning tasks~\cite{zheng2025diffusion, liao2024diffusiondrive}, employing techniques like truncated denoising and classifier guidance to generate closed-loop feasible plans with controllable behaviors. Their ability to incorporate external priors~\cite{lian2023llm} and scene-level context~\cite{zhong2023guided} further enhances their adaptability in generative planning pipelines.

Despite these strengths, diffusion models typically operate with implicit scene representations and lack semantic interpretability or high-level decision awareness~\cite{chen2024overview, yang2024diffusion}. Their reliance on low-level signals limits their ability to reason about abstract goals, safety constraints, or social norms, which is essential for human-aligned planning in urban scenarios. Moreover, their iterative sampling nature introduces significant inference latency, often requiring manually crafted guidance mechanisms to meet task-specific requirements~\cite{yang2024diffusion}. In practice, dozens of denoising steps are often needed to generate high-quality trajectories, posing a substantial computational burden~\cite{mao2023leapfrog, li2025predictive}.

Our proposed KnowDiffuser addresses these challenges by explicitly initializing the diffusion process with semantically meaningful prior-trajectories derived from high-level reasoning. This structured initialization, coupled with a two-stage denoising refinement, enables interpretable and semantically aligned trajectory generation while preserving the generative flexibility of diffusion models.

\subsection{Language Models in Autonomous Driving}

Language Models (LMs) have shown significant potential in enhancing the reasoning and contextual understanding of AD systems. These models offer human-like semantic comprehension, commonsense reasoning, and situational adaptability for high-level decision-making~\cite{xu2024drivegpt4, tian2024drivevlm, wen2023dilu, cui2024llm4ad}. Recent work has explored incorporating LMs into autonomous driving through hierarchical planning~\cite{wang2023drivemlm, long2024vlm, jiang2024senna}, interpretable behavior generation~\cite{zhang2024dima}, and world modeling~\cite{zhao2024drivedreamer2, gao2024vista}, leading to more robust and explainable control strategies. Surveys have highlighted the potential of LMs to replace fragile rule-based systems with more adaptable, knowledge-driven reasoning engines~\cite{cui2024survey, yang2023llm4drive}.

However, LMs operate over discrete token spaces and struggle to produce the continuous, physically constrained control signals required for real-world trajectory execution~\cite{xing2025openemma}. While they perform well in high-level semantic tasks, they are fundamentally unsuited for generating low-level motion plans. Existing integration approaches, such as LM-guided reward shaping or text-conditioned planning~\cite{zhong2023languageguided}, remain indirect and often lack explicit semantic grounding in the resulting trajectories.

To bridge this gap, KnowDiffuser establishes a tight integration between knowledge-driven high-level reasoning and diffusion-based low-level planning. Specifically, it uses LMs to output meta-actions rather than trajectories, which are then used to generate executable and high-quality motion plans through a diffusion model. This principled coupling bridges the semantic-physical divide, enabling a new class of interpretable and knowledge-informed planners for AD.

\section{Methodology}

\subsection{Overview}
\begin{figure*}[!h]
    \centering
    \includegraphics[width=0.8\textwidth]{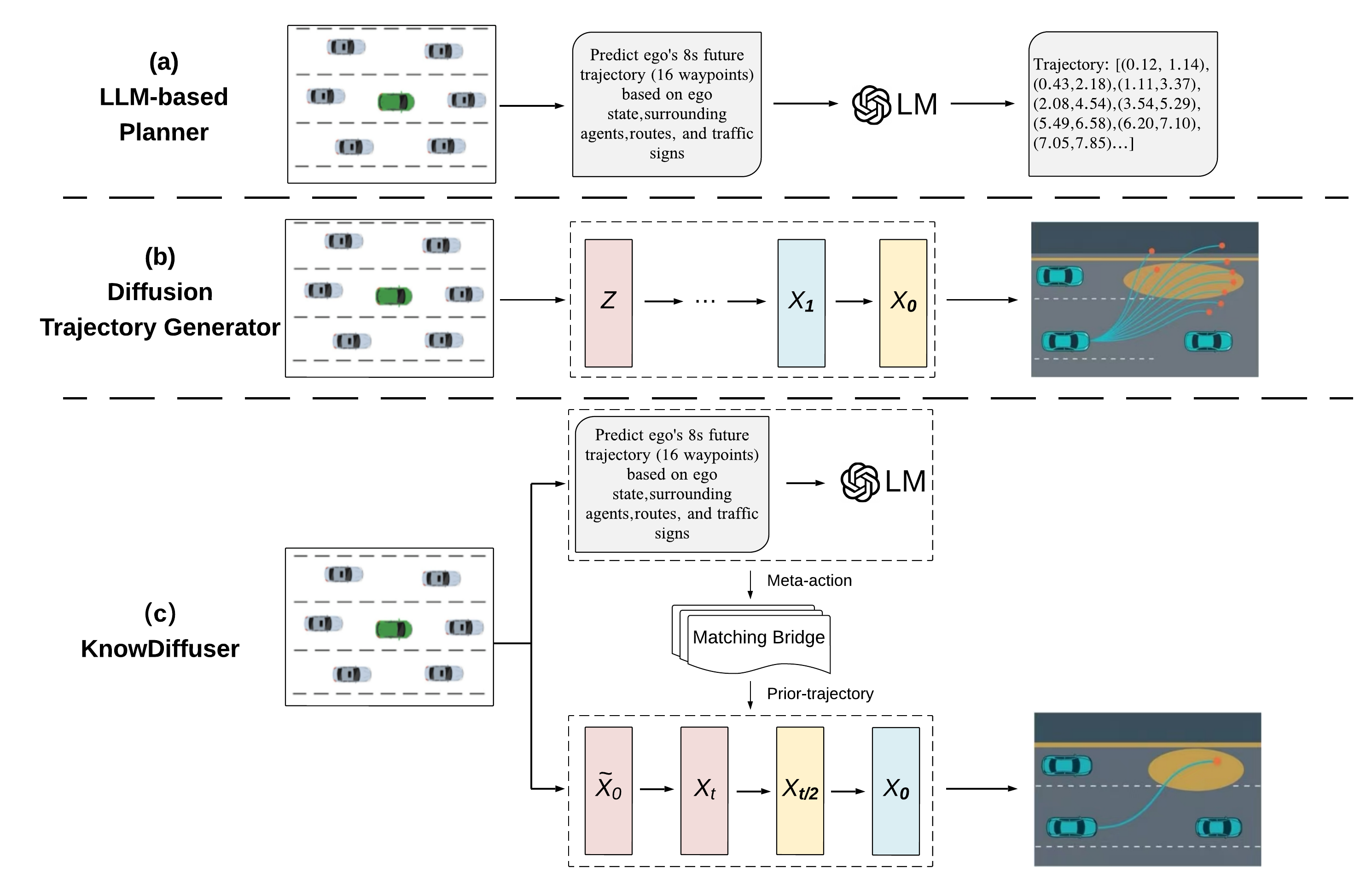}
    \caption{
        Comparative Overview of Planning Frameworks. 
        (a) LM-based Planner: Predicts future ego trajectories via semantic reasoning, offering interpretability but lacking in low-level control precision. 
        (b) Diffusion Trajectory Generator: Generates physically plausible trajectories through iterative denoising but lack of semantic understanding. 
        (c) KnowDiffuser: Combines LM-driven high-level meta-action planning with prior-trajectory diffusion two-step denoising, enabling semantically grounded yet physically feasible motion generation.
    }
    \label{fig:knowdiffuser_framework}
\end{figure*}

We propose KnowDiffuser, a knowledge-guided motion planning framework that tightly integrates a high-level decision module powered by an LM with a low-level diffusion-based trajectory generation module, as illustrated in Fig.~\ref{fig:knowdiffuser_framework}. The core idea is to bridge the semantic reasoning capabilities of LMs and the physical feasibility and diversity of diffusion models through a structured, multi-stage pipeline consisting of four key components:

First, we construct a meta-action to prior-trajectory matching library by analyzing large-scale driving logs. Each meta-action (e.g., “go straight”, “turn left/right”, “stop”) is associated with a representative prior trajectory derived from clustered historical driving segments. This library serves as a semantic grounding space that enables mapping discrete high-level intentions to continuous motion templates. Second, the high-level decision module utilizes an LM to reason over structured scene representations, including ego-vehicle state, surrounding agents, road topology, and traffic signals. The LM outputs a discrete meta-action that captures the intended behavior in context. Third, a meta-action to prior-trajectory matching bridge maps the selected meta-action to its corresponding prior-trajectory from the prebuilt library. This acts as a semantic-to-physical bridge, providing a behaviorally aligned initialization for the downstream generation process. Last, the low-level diffusion trajectory generator initializes from the retrieved prior-trajectory and refines it through a truncated two-stage denoising process. Instead of starting from pure Gaussian noise, the model injects mild noise into the prior trajectory, enabling faster inference while preserving semantic intent and physical feasibility. This design significantly enhances both planning efficiency and trajectory realism.

Together, these components form an intelligent, interpretable, and physically grounded planning pipeline that enables real-time, intention-aware motion generation in complex driving environments.

\subsection{Meta-Action and Prior-Trajectory Matching Library Construction}

As the first stage of our planning framework, we construct a structured library that maps discrete \textit{meta-actions} to their corresponding representative \textit{prior-trajectories}. This library provides a compact, interpretable, and physically grounded set of motion templates extracted from real-world driving data, forming the basis for downstream trajectory generation.

We begin by segmenting nuPlan \cite{caesar2021nuplan} driving logs into 8-second temporal windows, each capturing a coherent and self-contained driving maneuver. This duration is empirically validated to balance completeness and segmentation granularity~\cite{zheng2025diffusion}. For each segment, we extract the ego-centric trajectory and encode it as a sequence of 2D waypoints over a fixed horizon \( H \):
\begin{equation}
\tau = \{(x_1, y_1), (x_2, y_2), \ldots, (x_H, y_H)\}, \quad \tau \in \mathbb{R}^{H \times 2}
\end{equation}

To characterize the dynamics of each trajectory, we compute two summary features:
\begin{align}
\bar{v} &= \frac{1}{H - 1} \sum_{i=1}^{H-1} \frac{d\big((x_i, y_i), (x_{i+1}, y_{i+1})\big)}{\Delta t}, \\
\Delta \theta &= \sum_{i=2}^{H-1} |\theta_i - \theta_{i-1}|
\end{align}
where \( \bar{v} \) is the average speed and \( \Delta \theta \) is the total heading variation computed via atan2-based orientation. In addition, we compute the average longitudinal acceleration \( \bar{a} \) over the trajectory to capture speed variation. We define the trajectory feature vector as:
\begin{equation}
\phi(\tau) = \left[ \bar{v}, \bar{a}, \Delta \theta \right]
\end{equation}

We adopt a rule-based classification scheme to ensure semantic interpretability. Each 8-second trajectory is classified using predefined geometric and kinematic thresholds that capture both lateral displacement, heading change, and acceleration characteristics. Specifically, trajectories are categorized into canonical maneuver types including lane keeping (\(\Delta y < 0.5\,\text{m}\)), lane change (\(2.0\,\text{m} < \Delta y < 4.0\,\text{m}\)), and turning (\(\Delta \psi > 15^\circ\)). In addition, motion dynamics are characterized by acceleration or deceleration (\(\bar{a} > 0.5\,\text{m/s}^2\) or \(\bar{a} < -0.5\,\text{m/s}^2\)) and speed keeping (\(|\bar{a}| < 0.3\,\text{m/s}^2\)). The Cartesian combination of maneuver geometry and speed profiles defines the meta-action set \( \mathcal{A}_{\text{meta}} \).

Each trajectory is deterministically assigned a meta-action label:
\begin{equation}
a_j = \mathcal{M}_{\text{rule}}(\phi(\tau_j)), \quad a_j \in \mathcal{A}_{\text{meta}}
\end{equation}

Trajectories with the same meta-action label are grouped, and the corresponding \textit{prior-trajectory} \( \hat{\tau}_j \) is computed by time-wise averaging over the spatial coordinates:
\[
\hat{\tau}_j = \text{mean}\left(\{\tau_i \mid a_i = a_j\}\right)
\]

Each pair \( (a_j, \hat{\tau}_j) \) defines an entry in the library, ensuring that the mapping \( a_j \leftrightarrow \hat{\tau}_j \) is bijective and indexable. This results in a compact and interpretable trajectory library that captures canonical motion patterns grounded in real-world data, serving as the foundation for structured and behaviorally aligned planning.

\subsection{High-level Decision-making Module}

The high-level decision-making module is responsible for interpreting complex driving scenarios and generating a discrete \textit{meta-action} that captures the intended driving behavior. At its core is an LM, which performs context-aware reasoning over a structured understanding of traffic scenes from multi-source observations. Scene observations encode four key components of the traffic understanding: (1) the ego vehicle's state, including position, velocity, and heading; (2) the dynamic attributes of surrounding agents, such as type, trajectory, and proximity; (3) the road topology, including lane connectivity, intersections, and routing constraints; and (4) traffic control elements, such as traffic lights and stop signs. 

Formally, let the high-level observation at time \( t \) be denoted as:
\begin{equation}
\mathcal{O}_t = \{\mathbf{x}_t^{ego}, \mathcal{A}_t, \mathcal{M}_t, \mathcal{T}_t\}
\end{equation}
where \( \mathbf{x}_t^{ego} \) is the ego vehicle's state, \( \mathcal{A}_t \) is the set of surrounding agents, \( \mathcal{M}_t \) encodes the map topology, and \( \mathcal{T}_t \) represents traffic signals.

The observation is then encoded into a structured prompt:
\begin{equation}
\mathcal{P}_t = \mathcal{E}_{\text{prompt}}(\mathcal{O}_t)
\end{equation}

This prompt \( \mathcal{P}_t \) is fed into an LM \( \mathcal{L} \), which outputs a discrete high-level decision in the form of a meta-action:
\begin{equation}
a_t = \mathcal{L}(\mathcal{P}_t), \quad a_t \in \mathcal{A}_{\text{meta}}
\end{equation}

The meta-action set \( \mathcal{A}_{\text{meta}} \) is defined by composing interpretable driving intents along two behavioral axes: velocity and direction. The velocity axis includes \textit{accelerate}, \textit{cruise}, \textit{decelerate}, and \textit{brake}; the directional axis includes \textit{go straight}, \textit{left turn}, \textit{right turn}, \textit{U-turn}. These combinations define a compact and semantically meaningful set of driving behaviors.

By leveraging scene observations and commonsense reasoning, the LM interprets complex urban scenarios and outputs an intention-aware meta-action. This high-level decision serves as the behavioral directive for subsequent trajectory planning stages.

\subsection{Meta-Action to Prior-Trajectory Matching Bridge Mechanism}

To effectively align high-level semantic intent with low-level continuous trajectory generation, we introduce a bridge module that maps discrete \textit{meta-actions} to their corresponding representative \textit{prior-trajectories}. This bridging mechanism ensures that the abstract decisions made by the LM are grounded in physically plausible motion patterns extracted from real-world data.

Given a meta-action \( a_t \in \mathcal{A}_{\text{meta}} \) predicted by the high-level decision module at time \( t \), the bridge module performs a deterministic retrieval from a preconstructed library of indexed trajectory templates. Formally, we define this retrieval process as:
\begin{equation}
\hat{\tau}_t = \mathcal{E}_{\text{lookup}}(a_t), \quad \hat{\tau}_t \in \mathbb{R}^{H \times 2}
\end{equation}
where \( \mathcal{E}_{\text{lookup}}: \mathcal{A}_{\text{meta}} \rightarrow \mathbb{R}^{H \times 2} \) is a bijective mapping from each semantic meta-action to a trajectory prototype of fixed horizon \( H \).

The trajectory library used in this mapping is constructed offline from large-scale driving logs, as described in the meta-action to prior-trajectory library section. Each entry \( (a_j, \hat{\tau}_j) \) in the library corresponds to a statistically averaged trajectory that reflects a canonical behavior. 

This bridging process serves two critical roles. First, it translates high-level intent into a physically grounded initialization for trajectory generation. Second, it imposes a structured prior that constrains the sampling space of the diffusion model, thereby enhancing semantic alignment and improving generation efficiency.

The resulting prior-trajectory \( \hat{\tau}_t \) serves as the anchor for the subsequent diffusion-based refinement process. By indexing from a discrete semantic space into a continuous motion manifold, this bridge establishes a coherent interface between semantic reasoning and physical planning.

\subsection{Low-Level Trajectory Generation Module}

The low-level planner is responsible for generating high-resolution, physically feasible motion trajectories conditioned on the high-level semantic intent. Rather than initializing from pure Gaussian noise, our diffusion-based planner starts from a structured \textit{prior-trajectory} \( \tilde{x}_0 \in \mathbb{R}^{T \times 2} \) retrieved via meta-action matching. This prior encodes a coarse but semantically aligned motion pattern, serving as an informative initialization that guides the trajectory refinement process.

To generate a diverse yet behaviorally consistent trajectory, we introduce a \textbf{truncated two-step denoising mechanism}. Specifically, the prior trajectory is first perturbed with small amounts of Gaussian noise at two successive diffusion timesteps to simulate uncertainty while avoiding full noising. This results in a softly corrupted anchor trajectory that retains structural intent yet allows flexibility in the final output. The resulting noisy input is then fed into a transformer-based DiT decoder, which performs a learned denoising process conditioned on route context and current vehicle state.

This design achieves several objectives: (1) it preserves high-level behavioral consistency by anchoring generation to the prior; (2) it improves sampling efficiency by bypassing early diffusion steps; and (3) it produces diverse, physically grounded trajectories aligned with both semantic reasoning and vehicle dynamics. The entire generation process is trained end-to-end using a standard diffusion loss that reconstructs the ground-truth trajectory from noise-corrupted inputs.

\paragraph{Training Phase.}

During training, as shown in Algorithm~\ref{alg:training}, the diffusion model is trained to reconstruct the ground-truth future trajectory \( x_0 \in \mathbb{R}^{T \times 4} \), where each frame includes the agent's spatial position and heading vector. Instead of directly generating the trajectory, the model learns to reverse a stochastic corruption process based on a Variance Preserving Stochastic Differential Equation (VPSDE), which gradually perturbs the clean trajectory into noise.

\begin{algorithm}[H]
\caption{Training of Diffusion Predictor}
\label{alg:training}
\begin{algorithmic}[1]
\STATE \textbf{Input:} Ground-truth trajectory $x_0 \in \mathbb{R}^{T \times 4}$, current state $s \in \mathbb{R}^{1 \times 4}$, timestep sampler $t \sim \mathcal{U}(\varepsilon, 1)$
\STATE \textbf{Output:} Training loss $\mathcal{L}_{\text{diff}}$
\STATE Sample $t \sim \mathcal{U}(\varepsilon, 1)$
\STATE Sample $\epsilon \sim \mathcal{N}(0, \mathbf{I})$
\STATE $x_t \leftarrow \sqrt{\bar{\alpha}_t} \cdot x_0 + \sqrt{1 - \bar{\alpha}_t} \cdot \epsilon$
\STATE $x_{\text{input}} \leftarrow \texttt{concat}(s, x_t)$
\STATE $\hat{x}_0 \leftarrow \texttt{DiT}(x_{\text{input}}, t, \text{context})$
\STATE $\mathcal{L}_{\text{diff}} \leftarrow \left\| \hat{x}_0 - x_0 \right\|^2$
\STATE \textbf{Return:} $\mathcal{L}_{\text{diff}}$
\end{algorithmic}
\end{algorithm}

Concretely, given a randomly sampled timestep \( t \sim \mathcal{U}(\varepsilon, 1) \), where \( \varepsilon \) is a small positive constant, the clean trajectory is transformed into a noisy version using the closed-form marginal of the forward SDE:
\begin{equation}
x_t = \mu(t) + \sigma(t) \cdot \epsilon,\quad \epsilon \sim \mathcal{N}(0, \mathbf{I})
\end{equation}
where \( \mu(t) \) and \( \sigma(t) \) denote the time-dependent mean and standard deviation functions derived from the VPSDE. This formulation allows efficient sampling at arbitrary timesteps without requiring the full sequential simulation of diffusion steps.

The noisy trajectory \( x_t \in \mathbb{R}^{T \times 4} \) is then concatenated with the agent's current state \( x_{\text{cur}} \in \mathbb{R}^{1 \times 4} \), resulting in the model input \( x_{\text{input}} \in \mathbb{R}^{(T+1) \times 4} \). This input is passed to a DiT (Diffusion Transformer) decoder conditioned on the timestep embedding and scene context to produce a denoised prediction \( \hat{x}_0 \in \mathbb{R}^{T \times 4} \).

The training objective follows the standard \( x_0 \)-prediction loss, also known as the \texttt{x\_start} formulation, which directly minimizes the mean squared error between the predicted and true future trajectories:
\begin{equation}
\mathcal{L}_{\text{diff}} = \mathbb{E}_{x_0, t, \epsilon} \left[ \left\| \hat{x}_0 - x_0 \right\|^2 \right]
\end{equation}

This loss encourages the model to generate physically consistent and temporally smooth trajectories even under stochastic perturbations. By sampling over a range of noise levels during training, the model learns to handle varying degrees of uncertainty, enabling it to generalize robustly during inference and reconstruct high-fidelity, multimodal motion plans from noisy inputs.

\paragraph{Inference Phase.}
At inference time, as shown in Algorithm~\ref{alg:inference}, the model avoids the computational burden of simulating the full forward diffusion process from pure Gaussian noise. Instead, it leverages a behaviorally aligned prior trajectory \( \tilde{x}_0 \in \mathbb{R}^{T \times 2} \), which is deterministically retrieved based on the high-level meta-action predicted by the language model. This prior serves as a structured anchor that embeds semantic intent and physical plausibility into the generation process.

\begin{algorithm}[H]
\caption{Inference of Diffusion Predictor}
\label{alg:inference}
\begin{algorithmic}[1]
\STATE \textbf{Input:} Prior trajectory $\tilde{x}_0$, current state $s$, noise levels $\sigma(t_1), \sigma(t_2)$
\STATE \textbf{Output:} Predicted trajectory $\hat{\tau}$
\STATE $\tilde{x}_{\text{full}} \leftarrow \texttt{concat}(\tilde{x}_0, [1, 0])$
\STATE $\tilde{x}_{\text{ego}} \leftarrow \tilde{x}_{\text{full}}.\texttt{unsqueeze}(1)$
\STATE $\epsilon_1, \epsilon_2 \sim \mathcal{N}(0, I)$
\STATE $x_{t_1} \leftarrow \tilde{x}_{\text{ego}} + \sigma(t_1) \cdot \epsilon_1$
\STATE $x_{t_2} \leftarrow x_{t_1} + \sigma(t_2) \cdot \epsilon_2$
\STATE $x_T \leftarrow \texttt{concat}(s, x_{t_2})$
\STATE $\hat{x}_0 \leftarrow \texttt{DiT}(x_T, t_2, \text{context})$
\STATE \textbf{Return:} $\hat{\tau} \leftarrow \hat{x}_0[:, :, 1:]$
\end{algorithmic}
\end{algorithm}

To simulate intermediate noise levels while maintaining structural consistency, we adopt a truncated two-step noise injection scheme. First, the 2D prior trajectory \( \tilde{x}_0 \) is extended to 4D by appending a canonical heading vector \([1, 0]\), resulting in \( \tilde{x}_{\text{full}} \in \mathbb{R}^{T \times 4} \). The trajectory is then reshaped into batch format \( \tilde{x}_{\text{ego}} \in \mathbb{R}^{1 \times T \times 4} \) to match model input specifications.

Two independent Gaussian noise vectors \( \epsilon_1, \epsilon_2 \sim \mathcal{N}(0, \mathbf{I}) \) are sampled, and noise is injected in two steps to simulate approximate marginal distributions at time \( t_1 \) and \( t_2 \):
\begin{align}
x_{t_1} &= \tilde{x}_{\text{ego}} + \sigma(t_1) \cdot \epsilon_1 \\
x_{t_2} &= x_{t_1} + \sigma(t_2) \cdot \epsilon_2
\end{align}
where \( \sigma(t) \) denotes the marginal standard deviation computed from the VPSDE at timestep \( t \). This approximation introduces structured stochasticity without corrupting the trajectory entirely, enabling more efficient and semantically consistent sampling.

The noisy trajectory \( x_{t_2} \) is then concatenated with the agent's current state \( s \in \mathbb{R}^{1 \times 4} \), forming the full model input:
\begin{equation}
x_T = \texttt{concat}(s, x_{t_2}) \in \mathbb{R}^{1 \times (T+1) \times 4}
\end{equation}
This input is flattened and passed to the DiT decoder, which performs denoising conditioned on the timestep embedding \( t_2 \), route-level context, and surrounding scene features. The model outputs a denoised trajectory prediction \( \hat{x}_0 \in \mathbb{R}^{T \times 4} \), representing the final motion plan:
\begin{equation}
\hat{x}_0 = \texttt{DiT}(x_T, t_2, \texttt{context})
\end{equation}

Finally, the predicted trajectory is extracted (excluding the current state) and returned:
\begin{equation}
\hat{\tau} = \hat{x}_0[:, :, 1:]
\end{equation}

This truncated inference strategy ensures low-latency and high-fidelity trajectory generation by combining semantic priors with partial diffusion, effectively balancing interpretability, efficiency, and physical realism.
In summary, our diffusion-based planner leverages high-level semantic priors, efficient truncated perturbation, and transformer-based denoising to produce high-quality, behaviorally aligned trajectories suitable for real-time autonomous driving.

\section{Experiment}
\subsection{Experiment Setup}
We conduct comprehensive evaluations of KnowDiffuser on the large-scale real-world autonomous planning benchmark nuPlan \cite{caesar2021nuplan} using GPT-4o (2024) as the reasoning LMM.
To construct the training and evaluation corpus, we segment the raw nuPlan logs into 10-second driving sequences, resulting in a dataset of approximately 50,000 samples. Each sample is formatted as a trajectory prediction task, where the model receives the first 2 seconds of input and is tasked with predicting the ego vehicle's future trajectory over the next 8 seconds, discretized at 0.5-second intervals into 16 trajectory points.

The input to the model includes a rich set of information: (1) ego vehicle state history (position, velocity, heading), (2) surrounding agent states, (3) map-based road information, including lane geometry, road types, and the ego vehicle's spatial context within the road network, and (4) nearby traffic signs and signals. All input features are temporally aligned and normalized for efficient model training.

We evaluate our method under both open-loop (non-interactive, offline) settings and closed-loop settings. In the open-loop setup, we follow standard trajectory prediction evaluation using Average Displacement Error (ADE) and Final Displacement Error (FDE) to assess prediction performance. In the closed-loop setup, we evaluate our planner on the Val-14 scenarios using the official simulator and follow its built-in metric standards to assess closed-loop performance.

\subsection{Evaluation}

\textbf{Baselines.} We compare our method against a diverse set of existing planners, covering rule-based and learning-based paradigms without refine. Each baseline represents a distinct design philosophy within the open-loop planning task. Implementation details and references are as follows:

\begin{itemize}
    \item \textbf{IDM} \cite{treiber2000congested}: A classical rule-based planner implemented by nuPlan. It serves as a simple yet widely used baseline for evaluating trajectory planning performance.

    \item \textbf{UrbanDriver} \cite{scheel2021urban}: A reinforcement learning-based method using policy gradient optimization, implemented by nuPlan to simulate intelligent behavior under interactive conditions.
    
    \item \textbf{PlanCNN} \cite{renz2022plant}: A learning-based planner utilizing convolutional neural networks to process rasterized scene representations. It represents early neural approaches to open-loop prediction.

    \item \textbf{GameFormer} \cite{huang2023gameformer}: A transformer-based planner modeling multi-agent interactions as a dynamic game, outperforming traditional learning methods in reactive closed-loop evaluations.
    
    \item \textbf{GC-PGP} \cite{hallgarten2023prediction}: A behavior cloning planner based on imitation of expert demonstrations, but shows reduced generalization under complex driving scenarios.

    \item \textbf{PDM-Open} \cite{dauner2023parting}: A learning-based variant of the PDM family that follows the reference centerline without applying closed-loop feedback or hybrid control.

    \item \textbf{CKS-124m/CKS-1.5b} \cite{sun2023large}: Two transformer-based generative planners of different model sizes (124M and 1.5B parameters), capable of producing multimodal trajectory distributions with high realism and diversity.
    
    \item \textbf{PlanTF} \cite{cheng2024rethinking}: A pure imitation learning planner that directly regresses future motion from state and map inputs, removing reliance on rule-based modules while maintaining high planning fidelity.
    
    \item \textbf{GUMP-m} \cite{hu2024solving}: A guided diffusion-based motion planner with multi-objective scoring. It integrates safety, comfort, and social-awareness criteria for trajectory selection, achieving strong performance in open-loop planning.
\end{itemize}

\begin{table}[ht]
\centering
\caption{Open-loop planning results on the nuPlan test set. Lower is better.}
\label{tab:open_loop}
\small
\resizebox{\linewidth}{!}{
\begin{tabular}{llccccc}
\toprule
\textbf{Type} & \textbf{Methods} & \textbf{8sADE}↓ & \textbf{3sFDE}↓ & \textbf{5sFDE}↓ & \textbf{8sFDE}↓ & \textbf{MR}↓ \\
\midrule
\multirow{1}{*}{Rule-based}
& IDM & 9.600 & 6.256 & 10.076 & 16.993 & 0.552 \\
\midrule
\multirow{7}{*}{Learning-based}
& PlanCNN       & 2.468 & 0.955 & 2.486 & 5.936 & 0.064 \\
& UrbanDriver   & 2.667 & 1.497 & 2.815 & 5.453 & 0.064 \\
& PDM-Open      & 2.375 & 0.715 & 2.060 & 5.296 & 0.042 \\
& CKS-124m      & 1.777 & 0.951 & 2.105 & 4.515 & 0.053 \\
& CKS-1.5b      & 1.783 & 0.971 & 2.140 & 4.460 & 0.047 \\
& GUMP-m        & 1.820 & 0.743 & 1.833 & 4.453 & 0.046 \\
& \textbf{KnowDiffuser (Ours)} & \textbf{0.298} & \textbf{0.245} & \textbf{0.337} & \textbf{0.568} & \textbf{0.021} \\
\bottomrule
\end{tabular}
}
\end{table}

\noindent\textbf{Open-loop Main Results.} Evaluation results on the nuPlan benchmark are presented in Table~\ref{tab:open_loop}, showing a quantitative comparison of open-loop planning performance on the nuPlan test set. Across all five evaluation metrics, our proposed \textbf{KnowDiffuser} significantly outperforms all baselines, including strong diffusion-based planners such as GUMP-m and transformer-based models like CKS-1.5b. Notably, KnowDiffuser achieves an exceptionally low 8sADE of \textbf{0.298} and 8sFDE of \textbf{0.568}, indicating that the generated trajectories maintain high accuracy even over long horizons. This marks a substantial improvement over GUMP-m (1.820 / 4.453) and CKS-124m (1.777 / 4.515), demonstrating KnowDiffuser's ability to model temporal consistency and trajectory fidelity.

In addition, our method reports the lowest Miss Rate (MR) of \textbf{0.021}, reflecting strong safety guarantees by minimizing planning failures where predicted trajectories deviate excessively from the ground truth. These improvements can be attributed to the integration of LM-guided high-level semantic priors and our structured two-step guided diffusion process, which jointly enhance both global intent alignment and low-level trajectory feasibility. The results validate that KnowDiffuser not only improves predictive precision but also offers robust and reliable planning suitable for deployment in real-world driving systems.

\noindent\textbf{Closed-loop Main Results.}
To rigorously assess the efficacy of our proposed method, KnowDiffuser, we conduct comprehensive evaluations on the nuPlan dataset under the closed-loop planning setting. Table \ref{tab:closed_loop_grouped} presents a quantitative comparison with a suite of state-of-the-art baselines, including both classical rule-based planners (IDM) and modern learning-based methods such as UrbanDriver, GC-PGP, PlanCNN, GameFormer, and PlanTF. The metrics include success rates on both non-reactive (NR) and reactive (R) settings across validation (Val) and test splits. Higher scores indicate better performance.

\begin{table}[ht]
\centering
\caption{Performance comparison of closed-loop planning on Val-14 of the nuPlan dataset. Higher is better.}
\label{tab:closed_loop_grouped}
\small
\resizebox{\linewidth}{!}{
\begin{tabular}{llcccc}
\toprule
\textbf{Type} & \textbf{Model} & \textbf{Val NR} & \textbf{Val R} & \textbf{Test NR} & \textbf{Test R} \\
\midrule
\multirow{1}{*}{Rule-based}
& IDM & 77.02 & 76.50 & - & - \\
\midrule
\multirow{5}{*}{Learning-based}
& UrbanDriver & 63.27 & 61.02 & 51.54 & 49.07 \\
& GC-PGP & 55.99 & 51.39 & 43.22 & 39.63 \\
& PlanCNN  & 72.89 & 71.75 & -  & - \\
& GameFormer & 80.80 & 79.31 & 66.59 & 68.83 \\
& PlanTF & 84.83 & 76.78 & 72.68 & 61.70 \\
& \textbf{KnowDiffuser (Ours)} & \textbf{87.50} & \textbf{81.25} & \textbf{86.94} & \textbf{81.10} \\
\bottomrule
\end{tabular}
}
\end{table}

Our method achieves significant performance improvements across all settings. On the validation set, KnowDiffuser attains a Val NR score of \textbf{87.50} and a Val R score of \textbf{81.25}, outperforming the strongest baseline, PlanTF, by +2.67 and +4.47 percentage points respectively. Similarly, in the test set where the generalization capability is more critically evaluated, KnowDiffuser achieves a Test NR of \textbf{86.94} and a Test R of \textbf{81.10}, substantially exceeding GameFormer by +20.35 and +12.27 points and PlanTF by +14.26 and +19.40 points in the corresponding categories.

These results decisively validate the core design philosophy of KnowDiffuser: integrating high-level semantic reasoning via language models with low-level, physically grounded trajectory generation through diffusion models. The notable improvements over both diffusion-based (PlanTF) and transformer-based (GameFormer) planners indicate that our two-stage truncated denoising process, coupled with meta-action-informed prior initialization, yields robust and interpretable trajectory plans. Moreover, the superior performance on the reactive (R) metrics underscores the method's adaptability to dynamic agent interactions and unexpected scene changes.

In conclusion, KnowDiffuser not only surpasses existing benchmarks on closed-loop planning, but also sets a new state-of-the-art in balancing semantic alignment, physical feasibility, and real-world generalization, demonstrating its strong potential for real-time autonomous driving applications.

\subsection{Ablation Study on LLM Scale}

To address concerns regarding the influence of language model scale on planning performance, we conduct an ablation study by replacing the reasoning LMM in our framework with smaller open-source models while keeping the diffusion-based trajectory planner unchanged. Specifically, we evaluate LLaMA-3B (small) and Qwen3-32B (medium) under the same experimental setup on the nuPlan benchmark.

Table~\ref{tab:llm_scale} summarizes the results. Both smaller models achieve noticeably lower performance compared to the large-scale LLM adopted in our framework (GPT-4o). In particular, LLaMA-3B obtains a score of 60.21, while Qwen3-32B achieves 65.32, both significantly below the performance achieved by GPT-4o.

\begin{table}[ht]
\centering
\caption{Ablation study on the effect of LLM scale. Higher is better.}
\label{tab:llm_scale}
\small
\begin{tabular}{lc}
\toprule
\textbf{LLM Backbone} & \textbf{Score} $\uparrow$ \\
\midrule
LLaMA-3B (Small) & 60.21 \\
Qwen3-32B (Medium) & 65.32 \\
GPT-4o (Large, Ours) & \textbf{81.10} \\
\bottomrule
\end{tabular}
\end{table}

Further analysis of the decision logs reveals that smaller models frequently collapse to repetitive meta-actions (e.g., ``stop'') in complex driving scenes. This behavior suggests that insufficient model capacity limits contextual understanding and reduces the reliability of high-level decision making. In contrast, the stronger reasoning capability of GPT-4o enables more consistent interpretation of scene semantics and more reliable meta-action selection, which ultimately leads to superior planning performance.

\section{Conclusion}

In this work, we proposed KnowDiffuser, a novel knowledge-guided motion planning framework that effectively bridges high-level semantic reasoning and low-level motion planning for AD. By integrating an LM for scene understanding and intent inference with a diffusion-based trajectory generator, KnowDiffuser achieves a unique synergy: interpretable high-level decisions are mapped to physically grounded motion plans through a structured prior initialization and an efficient two-stage denoising process.

Our extensive experiments on the nuPlan benchmark demonstrate that KnowDiffuser consistently outperforms state-of-the-art baselines across both open-loop and closed-loop settings. It achieves superior accuracy, robustness, and semantic alignment while significantly reducing inference latency, which is key to real-time deployment. These results validate the efficacy of combining language-driven decision-making with generative diffusion modeling, highlighting the importance of bridging the semantic-physical gap in AD.

In future work, we aim to enhance the LM component with Vision-Language Models to enable multimodal scene understanding and reasoning, allowing richer perception of complex driving environments. Additionally, we plan to integrate trajectory refinement models to further improve motion quality and safety, advancing toward more reliable and intelligent autonomous planning systems.

\bibliographystyle{IEEEtran}
\bibliography{reference}

@article{wen2023dilu,
  title={Dilu: A knowledge-driven approach to autonomous driving with large language models},
  author={Wen, Licheng and others},
  journal={arXiv preprint arXiv:2309.16292},
  year={2023}
}

@article{wang2023drivemlm,
  title={Drivemlm: Aligning multi-modal large language models with behavioral planning states for autonomous driving},
  author={Wang, Wenhai and others},
  journal={arXiv preprint arXiv:2312.09245},
  year={2023}
}

@inproceedings{cui2024survey,
  title={A survey on multimodal large language models for autonomous driving},
  author={Cui, Can and others},
  booktitle={Proceedings of the IEEE/CVF Winter Conference on Applications of Computer Vision},
  pages={958--979},
  year={2024}
}

@article{yang2023llm4drive,
  title={LLM4Drive: A Survey of Large Language Models for Autonomous Driving},
  author={Yang, Zhenjie and others},
  journal={arXiv e-prints arXiv:2311},
  year={2023}
}

@inproceedings{tian2024drivevlm,
  title={Drivevlm: The convergence of autonomous driving and large vision-language models},
  author={Tian, Xiaoyu and Gu, Junru and Li, Bailin and Liu, Yicheng and Hu, Chenxu and Wang, Yang and Zhan, Kun and Jia, Peng and Lang, Xianpeng and Zhao, Hang},
  booktitle={CoRL},
  year={2024}
}

@article{xu2024drivegpt4,
  title={Drivegpt4: Interpretable end-to-end autonomous driving via large language model},
  author={Xu, Zhenhua and Zhang, Yujia and Xie, Enze and Zhao, Zhen and Guo, Yong and Wong, Kwan-Yee K and Li, Zhenguo and Zhao, Hengshuang},
  journal={RA-L},
  year={2024},
}

@article{caesar2021nuplan,
  title={nuplan: A closed-loop ml-based planning benchmark for autonomous vehicles},
  author={Caesar, Holger and Kabzan, Juraj and Tan, Kok Seang and Fong, Whye Kit and Wolff, Eric and Lang, Alex and Fletcher, Luke and Beijbom, Oscar and Omari, Sammy},
  journal={arXiv preprint arXiv:2106.11810},
  year={2021}
}

@article{ho2020denoising,
  title={Denoising diffusion probabilistic models},
  author={Ho, Jonathan and Jain, Ajay and Abbeel, Pieter},
  journal={Advances in neural information processing systems},
  volume={33},
  pages={6840--6851},
  year={2020}
}

@inproceedings{sohl2015deep,
  title={Deep unsupervised learning using nonequilibrium thermodynamics},
  author={Sohl-Dickstein, Jascha and Weiss, Eric and Maheswaranathan, Niru and Ganguli, Surya},
  booktitle={International conference on machine learning},
  pages={2256--2265},
  year={2015},
  organization={pmlr}
}

@misc{scheel2021urban,
      title={Urban Driver: Learning to Drive from Real-world Demonstrations Using Policy Gradients}, 
      author={Oliver Scheel and Luca Bergamini and Maciej Wołczyk and Błażej Osiński and Peter Ondruska},
      year={2021},
      eprint={2109.13333},
      archivePrefix={arXiv},
      primaryClass={id='cs.RO' full_name='Robotics' is_active=True alt_name=None in_archive='cs' is_general=False description='Roughly includes material in ACM Subject Class I.2.9.'}
}

@misc{song2021scorebased,
      title={Score-Based Generative Modeling through Stochastic Differential Equations}, 
      author={Yang Song and Jascha Sohl-Dickstein and Diederik P. Kingma and Abhishek Kumar and Stefano Ermon and Ben Poole},
      year={2021},
      eprint={2011.13456},
      archivePrefix={arXiv},
      primaryClass={id='cs.LG' full_name='Machine Learning' is_active=True alt_name=None in_archive='cs' is_general=False description='Papers on all aspects of machine learning research (supervised, unsupervised, reinforcement learning, bandit problems, and so on) including also robustness, explanation, fairness, and methodology. cs.LG is also an appropriate primary category for applications of machine learning methods.'}
}

@inproceedings{jiang2023motiondiffuser,
  title={Motiondiffuser: Controllable multi-agent motion prediction using diffusion},
  author={Jiang, Chiyu and Cornman, Andre and Park, Cheolho and Sapp, Benjamin and Zhou, Yin and Anguelov, Dragomir and others},
  booktitle={Proceedings of the IEEE/CVF Conference on Computer Vision and Pattern Recognition},
  pages={9644--9653},
  year={2023}
}

@article{yang2024diffusion,
  title={Diffusion-es: Gradient-free planning with diffusion for autonomous driving and zero-shot instruction following},
  author={Yang, Brian and Su, Huangyuan and Gkanatsios, Nikolaos and Ke, Tsung-Wei and Jain, Ayush and Schneider, Jeff and Fragkiadaki, Katerina},
  journal={arXiv preprint arXiv:2402.06559},
  year={2024}
}

@article{treiber2000congested,
  title={Congested traffic states in empirical observations and microscopic simulations},
  author={Treiber, Martin and Hennecke, Ansgar and Helbing, Dirk},
  journal={Physical review E},
  volume={62},
  number={2},
  pages={1805},
  year={2000},
  publisher={APS}
}

@inproceedings{zhong2023guided,
  title={Guided conditional diffusion for controllable traffic simulation},
  author={Zhong, Ziyuan and Rempe, Davis and Xu, Danfei and Chen, Yuxiao and Veer, Sushant and Che, Tong and Ray, Baishakhi and Pavone, Marco},
  booktitle={2023 IEEE International Conference on Robotics and Automation (ICRA)},
  pages={3560--3566},
  year={2023},
  organization={IEEE}
}

@article{sun2023large,
  title={Large Trajectory Models are Scalable Motion Predictors and Planners},
  author={Sun, Qiao and Zhang, Shiduo and Ma, Danjiao and Shi, Jingzhe and Li, Derun and Luo, Simian and Wang, Yu and Xu, Ningyi and Cao, Guangzhi and Zhao, Hang},
  journal={arXiv preprint arXiv:2310.19620},
  year={2023}
}

@inproceedings{sharan2024plan,
  title={Plan diffuser: Grounding llm planners with diffusion models for robotic manipulation},
  author={Sharan, SP and Zhao, Ruihan and Wang, Zhangyang and Chinchali, Sandeep P and others},
  booktitle={Bridging the Gap between Cognitive Science and Robot Learning in the Real World: Progresses and New Directions},
  year={2024}
}

@inproceedings{mao2023leapfrog,
  title={Leapfrog diffusion model for stochastic trajectory prediction},
  author={Mao, Weibo and Xu, Chenxin and Zhu, Qi and Chen, Siheng and Wang, Yanfeng},
  booktitle={Proceedings of the IEEE/CVF conference on computer vision and pattern recognition},
  pages={5517--5526},
  year={2023}
}

@inproceedings{gu2022stochastic,
  title={Stochastic trajectory prediction via motion indeterminacy diffusion},
  author={Gu, Tianpei and Chen, Guangyi and Li, Junlong and Lin, Chunze and Rao, Yongming and Zhou, Jie and Lu, Jiwen},
  booktitle={Proceedings of the IEEE/CVF conference on computer vision and pattern recognition},
  pages={17113--17122},
  year={2022}
}

@article{zheng2025diffusion,
  title={Diffusion-Based Planning for Autonomous Driving with Flexible Guidance},
  author={Zheng, Yinan and Liang, Ruiming and Zheng, Kexin and Zheng, Jinliang and Mao, Liyuan and Li, Jianxiong and Gu, Weihao and Ai, Rui and Li, Shengbo Eben and Zhan, Xianyuan and others},
  journal={arXiv preprint arXiv:2501.15564},
  year={2025}
}

@article{liao2024diffusiondrive,
  title={DiffusionDrive: Truncated Diffusion Model for End-to-End Autonomous Driving},
  author={Liao, Bencheng and Chen, Shaoyu and Yin, Haoran and Jiang, Bo and Wang, Cheng and Yan, Sixu and Zhang, Xinbang and Li, Xiangyu and Zhang, Ying and Zhang, Qian and others},
  journal={arXiv preprint arXiv:2411.15139},
  year={2024}
}

@article{luo2024pkrd,
  title={Pkrd-cot: A unified chain-of-thought prompting for multi-modal large language models in autonomous driving},
  author={Luo, Xuewen and Ding, Fan and Song, Yinsheng and Zhang, Xiaofeng and Loo, Junnyong},
  journal={arXiv preprint arXiv:2412.02025},
  year={2024}
}

@article{long2024vlm,
  title={VLM-MPC: Vision Language Foundation Model (VLM)-Guided Model Predictive Controller (MPC) for Autonomous Driving},
  author={Long, Keke and Shi, Haotian and Liu, Jiaxi and Li, Xiaopeng},
  journal={arXiv preprint arXiv:2408.04821},
  year={2024}
}

@article{jiang2024senna,
  title={Senna: Bridging large vision-language models and end-to-end autonomous driving},
  author={Jiang, Bo and Chen, Shaoyu and Liao, Bencheng and Zhang, Xingyu and Yin, Wei and Zhang, Qian and Huang, Chang and Liu, Wenyu and Wang, Xinggang},
  journal={arXiv preprint arXiv:2410.22313},
  year={2024}
}

@article{chen2024overview,
  title={An overview of diffusion models: Applications, guided generation, statistical rates and optimization},
  author={Chen, Minshuo and Mei, Song and Fan, Jianqing and Wang, Mengdi},
  journal={arXiv preprint arXiv:2404.07771},
  year={2024}
}

@article{lian2023llm,
  title={Llm-grounded diffusion: Enhancing prompt understanding of text-to-image diffusion models with large language models},
  author={Lian, Long and Li, Boyi and Yala, Adam and Darrell, Trevor},
  journal={arXiv preprint arXiv:2305.13655},
  year={2023}
}

@inproceedings{xing2025openemma,
  title={Openemma: Open-source multimodal model for end-to-end autonomous driving},
  author={Xing, Shuo and Qian, Chengyuan and Wang, Yuping and Hua, Hongyuan and Tian, Kexin and Zhou, Yang and Tu, Zhengzhong},
  booktitle={Proceedings of the Winter Conference on Applications of Computer Vision},
  pages={1001--1009},
  year={2025}
}

@inproceedings{zhong2023languageguided,
  title = {Language-Guided Traffic Simulation via Scene-Level Diffusion},
  author = {Zhong, Ziyuan and Rempe, Davis and Chen, Yuxiao and Ivanovic, Boris and Cao, Yulong and Xu, Danfei and Pavone, Marco and Ray, Baishakhi},
  booktitle = {Proceedings of The 7th Conference on Robot Learning},
  pages = {144--177},
  year = {2023},
  editor = {Tan, Jie and Toussaint, Marc and Darvish, Kourosh},
  volume = {229},
  series = {Proceedings of Machine Learning Research},
  publisher = {PMLR}
}

@article{zhao2024drivedreamer2,
  title={DriveDreamer-2: LLM-Enhanced World Models for Diverse Driving Video Generation},
  author={Zhao, Guosheng and Wang, Xiaofeng and Zhu, Zheng and Chen, Xinze and Huang, Guan and Bao, Xiaoyi and Wang, Xingang},
  journal={arXiv preprint arXiv:2403.06845},
  year={2024}
}

@article{gao2024vista,
  title={Vista: A generalizable driving world model with high fidelity and versatile controllability},
  author={Gao, Shenyuan and Yang, Jiazhi and Chen, Li and Chitta, Kashyap and Qiu, Yihang and Geiger, Andreas and Zhang, Jun and Li, Hongyang},
  journal={arXiv preprint arXiv:2405.17398},
  year={2024}
}

@inproceedings{zhang2024dima,
  title={{DiMA}: Distilling Multi-modal Large Language Models for Efficient Visual Planning},
  author={Zhang, Yunchuan and Zhao, Zelin and Du, Yilun and Xie, Sirui and Zhou, Bolei},
  booktitle={Proceedings of the IEEE/CVF Conference on Computer Vision and Pattern Recognition (CVPR)},
  year={2024}
}

@misc{cui2024llm4ad,
      title={{LLM4AD}: A Systematic Benchmark for Evaluating the Instruction-Following Abilities of Large Language Models in Autonomous Driving},
      author={Yiming Cui and Zhipeng Cao and Zhepeng Wang and Zhijin Li and Wenjia Liu and Zeyu Li and Yueqi Duan and Ji Wu and Bolei Zhou},
      year={2024},
      eprint={2401.05316},
      archivePrefix={arXiv},
      primaryClass={cs.RO}
}

@article{renz2022plant,
  title={Plant: Explainable planning transformers via object-level representations},
  author={Renz, Katrin and Chitta, Kashyap and Mercea, Otniel-Bogdan and Koepke, A and Akata, Zeynep and Geiger, Andreas},
  journal={arXiv preprint arXiv:2210.14222},
  year={2022}
}

@inproceedings{dauner2023parting,
  title={Parting with misconceptions about learning-based vehicle motion planning},
  author={Dauner, Daniel and Hallgarten, Marcel and Geiger, Andreas and Chitta, Kashyap},
  booktitle={Conference on Robot Learning},
  pages={1268--1281},
  year={2023},
  organization={PMLR}
}

@inproceedings{hu2024solving,
  title={Solving motion planning tasks with a scalable generative model},
  author={Hu, Yihan and Chai, Siqi and Yang, Zhening and Qian, Jingyu and Li, Kun and Shao, Wenxin and Zhang, Haichao and Xu, Wei and Liu, Qiang},
  booktitle={European Conference on Computer Vision},
  pages={386--404},
  year={2024},
  organization={Springer}
}

@inproceedings{lin2025causal,
  title={Causal composition diffusion model for closed-loop traffic generation},
  author={Lin, Haohong and Huang, Xin and Phan, Tung and Hayden, David and Zhang, Huan and Zhao, Ding and Srinivasa, Siddhartha and Wolff, Eric and Chen, Hongge},
  booktitle={Proceedings of the Computer Vision and Pattern Recognition Conference},
  pages={27542--27552},
  year={2025}
}

@article{li2025predictive,
  title={Predictive Planner for Autonomous Driving with Consistency Models},
  author={Li, Anjian and Bae, Sangjae and Isele, David and Beeson, Ryne and Tariq, Faizan M},
  journal={arXiv preprint arXiv:2502.08033},
  year={2025}
}

@inproceedings{huang2023gameformer,
  title={Gameformer: Game-theoretic modeling and learning of transformer-based interactive prediction and planning for autonomous driving},
  author={Huang, Zhiyu and Liu, Haochen and Lv, Chen},
  booktitle={Proceedings of the IEEE/CVF International Conference on Computer Vision},
  pages={3903--3913},
  year={2023}
}

@inproceedings{cheng2024rethinking,
  title={Rethinking imitation-based planners for autonomous driving},
  author={Cheng, Jie and Chen, Yingbing and Mei, Xiaodong and Yang, Bowen and Li, Bo and Liu, Ming},
  booktitle={2024 IEEE International Conference on Robotics and Automation (ICRA)},
  pages={14123--14130},
  year={2024},
  organization={IEEE}
}

@inproceedings{hallgarten2023prediction,
  title={From prediction to planning with goal conditioned lane graph traversals},
  author={Hallgarten, Marcel and Stoll, Martin and Zell, Andreas},
  booktitle={2023 IEEE 26th International Conference on Intelligent Transportation Systems (ITSC)},
  pages={951--958},
  year={2023},
  organization={IEEE}
}


\end{document}